\documentclass[a4paper, 10pt, conference]{ieeeconf}      
\usepackage{graphicx}
\usepackage{amsfonts}
\usepackage{csquotes}
\usepackage{subfig}
\usepackage{url}
\usepackage{booktabs}
\usepackage[keeplastbox,spread,noautobase]{flushend}

\IEEEoverridecommandlockouts                              

\title{\LARGE \bf
Practical Aspects of Autonomous Exploration with a Kinect2 sensor
}

\author{Miroslav Kulich, Vojt\v{e}ch Lhotsk\'y, Libor P\v{r}eu\v{c}il
\thanks{The authors are with Czech Institute of Informatics, Robotics and Cybernetics, Czech Technical University in Prague,%
      Zikova 1903/4, 166 36 Prague, Czech Republic 
   {\tt\small kulich@cvut.cz,lhotsvoj@fel.cvut.cz, preucil@cvut.cz}%
}
}

\begin{document}

\maketitle
\thispagestyle{empty}
\pagestyle{empty}

\begin{abstract}
Exploration of an unknown environment by a mobile robot is a complex task involving solution of many fundamental problems from data processing, localization to high-level planning and decision making.
The exploration framework we developed is based on processing of RGBD data provided by a MS Kinect2 sensor, which allows to take advantage of state-of-the-art SLAM (Simultaneous Localization and Mapping) algorithms and to autonomously build a realistic 3D map of the environment with projected visual information about the scene.
In this paper, we describe practical issues that appeared during deployment of the framework in real indoor and outdoor environments and discuss especially properties of SLAM algorithms processing MS Kinect2 data on an embedded computer.
\end{abstract}

\section{Introduction}
Exploration is a process of autonomous navigation of a mobile robot in a priory unknown environment in order to build a map of this environment. 
An exploration algorithm is an iterative procedure consisting of map updating from current sensory data, determination of a new goal and  navigation towards this goal. Such an algorithm is terminated whenever a complete map of the environment is built. 

The used exploration approach~\cite{eapd} assumes that the robot moves on a plane and thus exploration can be done in 2D. Specifically, a polygonal map is used for goal determination and planning: a 2D scan is simulated from a single measurement of the MS Kinect2 and approximated with a polygon and particular polygons are then combined together by a modified Vatti algorithm~\cite{icaps}. 
To provide a correct pose of the robot, RTAB-Map~\cite{rtabmap},  a RGBD Graph-Based SLAM approach based on an incremental appearance-based loop closure detector is employed. This approach, moreover, builds a colored three dimensional model of the environment, which is a final product of the exploration.
RTAB-Map was chosen based on comparison of available SLAM libraries, see Table~\ref{tab:Comparison-of-SLAMs}.

The software part of the system has been realized as a set of ROS (Robot Operating System) nodes. 
Reading data from the Kinect sensor is done through the {\tt libfreenect2} library~\cite{libfreenect}, which provides basic drivers for Kinect2, while connection to ROS is provided by {\tt Kinect2 Bridge library}~\cite{bridge}. A modified version of ER1 robot by Evolution Robotics was used as a robotic platform for experiments and all computations were done on an Intel NUC mini computer with a Core i5 processor placed directly on the robot.

\section{Real environment issues}
\label{sec:real-env-issues}

\begin{figure*}
\subfloat[][]{\includegraphics[height=0.15\textheight]{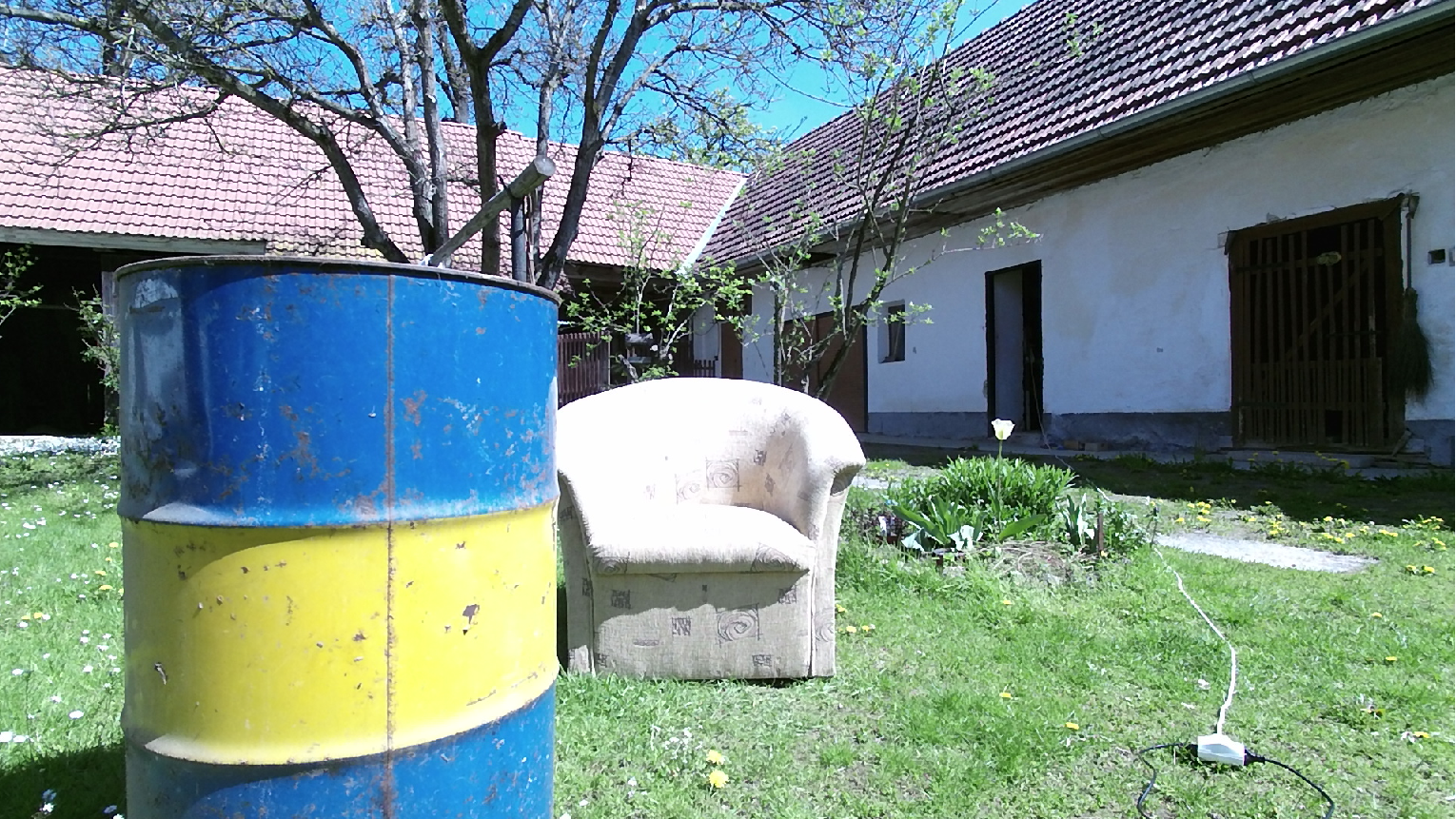}\label{fig:yard_rgb}}\hfill
\subfloat[][]{\includegraphics[height=0.15\textheight]{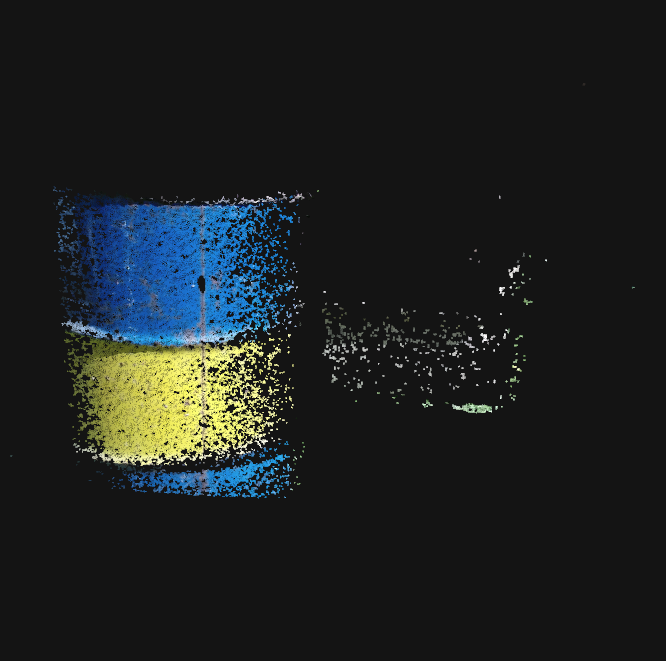}\label{fig:yard_sunny}}\hfill
\subfloat[][]{\includegraphics[height=0.15\textheight]{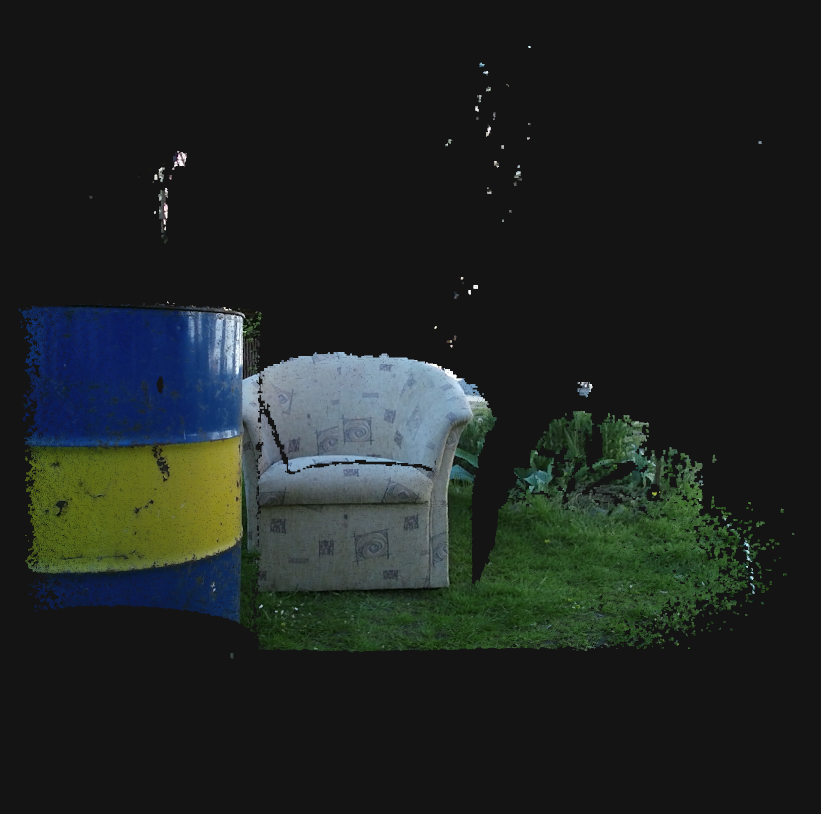}\label{fig:yard_shadow}}\hfill
\subfloat[][]{\includegraphics[height=0.15\textheight]{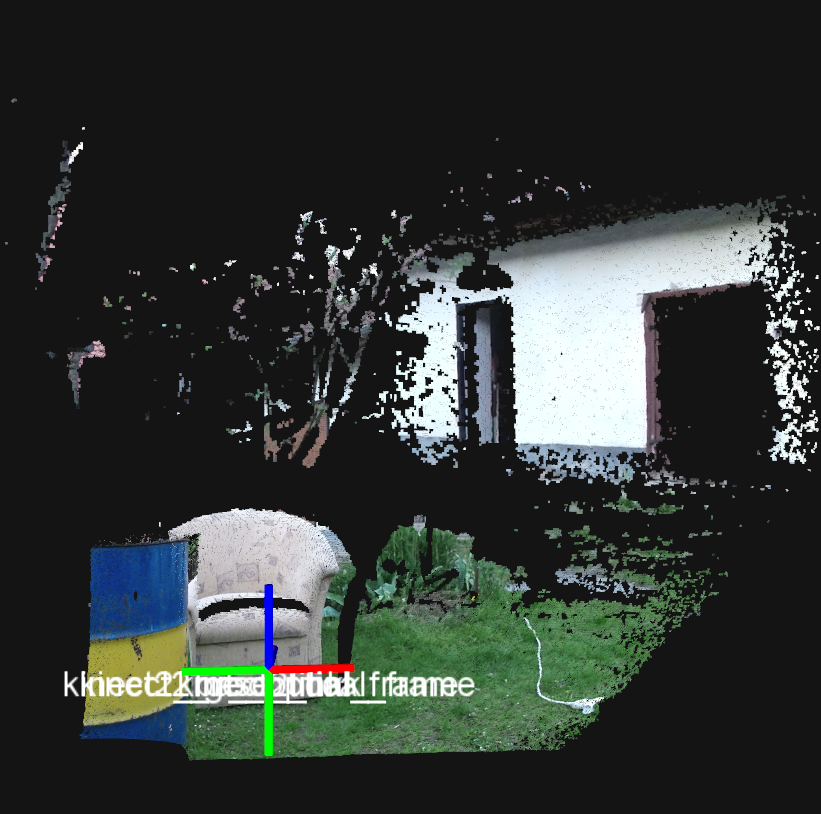}\label{fig:yard_evening}}\hfill
\caption{ (a) The reference environment, (b) Kinect 2 data in a~sunny day, (c) data with most of the scene covered in a~shadow (d) data in the evening}
\end{figure*}

Kinect 2 uses a~time of flight method for building a depth image and thus it should work (contrary to the first version) outdoor even in a~bright day.
An experiment was done in an outdoor environment depicted on Fig. \ref{fig:yard_rgb} to verify this assumption and \textbf{resistance} of the sensor \textbf{to sunlight}. 
The environment contains several objects at various distances from the sensor: a~barrel (1~m from the sensor), a~chair (2~m), flowers (3~m), and several trees and other objects placed 4 -- 15~m from the sensor.

\begin{figure}
  \begin{centering}
    \includegraphics[width=\columnwidth]{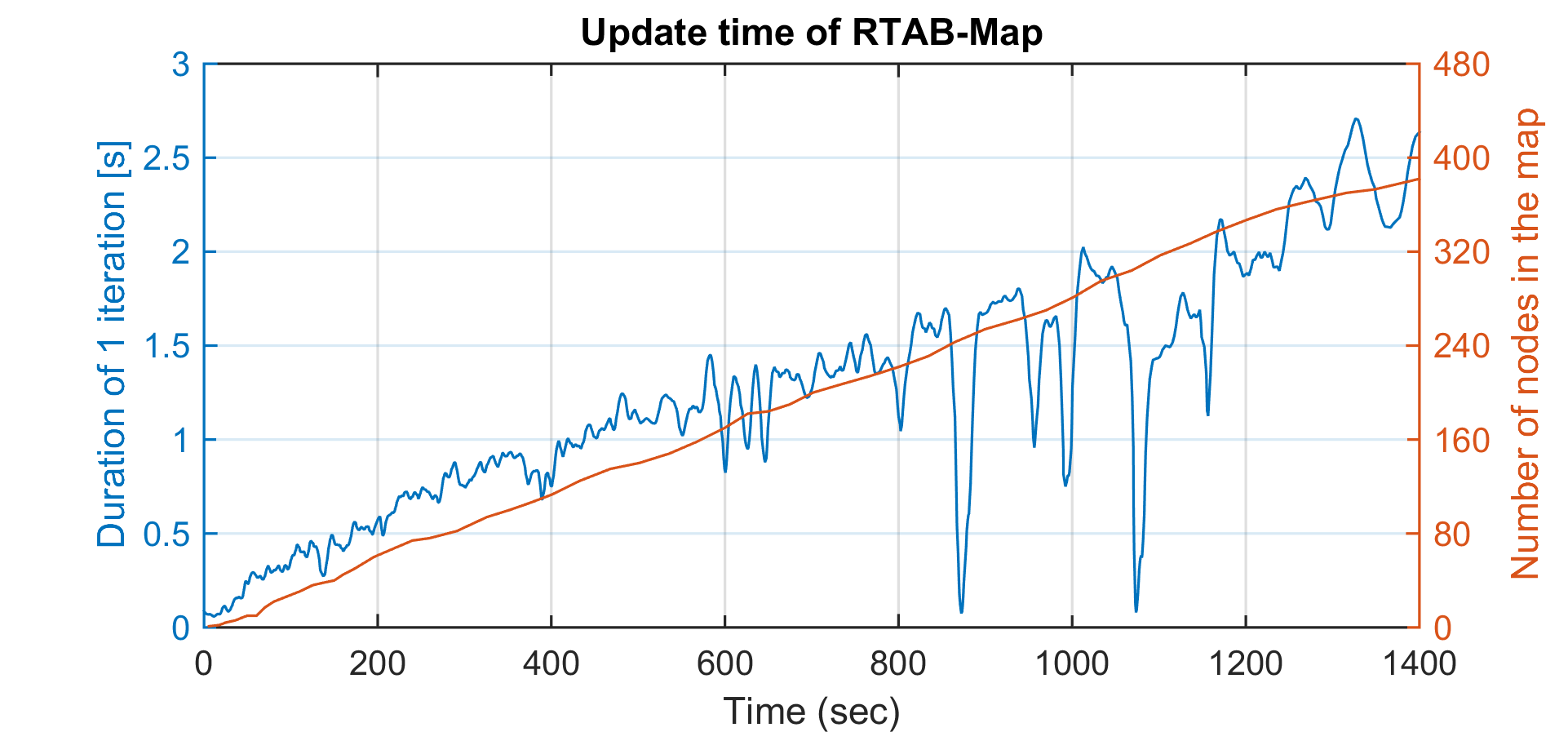}\caption[Update time of RTAB-Map]{\label{fig:update-time-rtab}
      The time of the RTAB-Map's update and the number of nodes in the  map depending on the length of time the mapping runs (test done on 
      Intel NUC computer).}
    
    \par\end{centering}
\end{figure}

The first experiment was done in a~very bright sunlight at 1 PM (May 7, 2016, near Prague). 
The results (Fig. \ref{fig:yard_sunny}) show that the data are very noisy and only the 
barrel is represented in an acceptable quality. 
The front part of the chair is still visible, but with a~large amount of missing points. 
Generally, the range of the Kinect 2 sensor in a~very bright day is approximately 1.2 -- 2.4~m depending on the material of the detected obstacle and many other factors (whether a~part of that obstacle is in a shadow, whether the sun shines directly on the sensor and on reflectivity of object's surface). Running RTAB-Map is thus nearly impossible as it can not  effectively localize due to the limited range and large amount of noise.

\begin{table}[h!]
\centering{}%
\begin{tabular}{p{.28\columnwidth} p{.28\columnwidth} p{.28\columnwidth}}


\toprule 
Rgbdslam v2~\cite{rgbd-mapping}  & RTAB-Map~\cite{rtabmap} & ORB-SLAM 2~\cite{orb-slam}\tabularnewline
\midrule[\heavyrulewidth] 
- Very slow updates of the map & Faster than Rgbdslam, slower than ORB-SLAM but provides a~fast visual odometry & + Very fast updates of the map\tabularnewline
\midrule
+ Can re-localize after looking at known position & Problems with re-localization, but it can use the wheel odometry & + Can successfully re-localize after being lost for a~long time, but
also becomes lost more often than the others\tabularnewline
\midrule
+ Well readable map, also published through ROS & + Well readable map, also published through ROS & - The map is not very readable (for people)\tabularnewline
\midrule
+ All important data is published through ROS topics & + All important data is published through ROS topics \& services & - No data is published through ROS topics\tabularnewline
\midrule
 & + Can utilize the odometry provided by the ER1 robot & \tabularnewline
\midrule
+ Wide range of parameters & + Wide range of parameters & - Low range of parameters\tabularnewline
\bottomrule 
\end{tabular}
\caption{\label{tab:Comparison-of-SLAMs}Comparison of RGBD SLAM libraries.}
\vspace{-1.5em}
\end{table}

The same area is also scanned later afternoon when the most of the scene is covered in shadow from the~building, but the sunlight is still bright so even areas covered in shadow are illuminated, see Fig. \ref{fig:yard_shadow}.
The range increased to 3.5 -- 6~m under these circumstances and the majority of objects within 3m range is well represented with almost no noise or outliers. 
RTAB-Map SLAM is able to run, but localization precision is worse than in ideal conditions.

The last experiment was done in the evening when the scene illumination is much lower, see Fig.~\ref{fig:yard_evening}).
The sensor correctly detected objects in 12~m range and the RTAB-Map worked with similar precision as indoors.

\textbf{The Field of view} of the sensor is only $70$~degrees, which causes big problems for the SLAM especially when the robot is rotating. 
Unrecoverable errors appear even though the angular velocity is limited to $17^\circ\cdot s^{-1}$ (lowering the velocity even more would make the exploration too slow).

When exploration runs for a~long period of time, its refresh duration is longer than 2~seconds (see the next paragraph) and the robot rotates by up to $34^\circ$. 
Due to the low field of view the SLAM can only match the features from roughly $50\%$ of the image.

RTAB-Map has an option for limiting the computation time, but at the cost of lower quality and even then the duration of the update gradually increases as the map grows. 
The SLAM must therefore rely on the data from the odometry.
Visual odometry in RTAB-Map is fast a precise enough to cover long-standing updates of RTAB-Map in most cases.
On the other hand, it can fail due to a low number of features and thus the only working source of the position data is wheel odometry, which slowly drifts.

\textbf{Duration of the RTAB-Map's update} is directly linked to the problem with a field of view. 
The time of the update gradually increases. 
The update time was set to $0.5$~s in the experiment presented in Fig.~\ref{fig:update-time-rtab}. 
As can be seen, this threshold is exceeded in approximately $200$~s after the start and the time of the update continues to grow as the map expands.

\textbf{Reflective surfaces} in the environment cause large amount of noise and also silhouettes of the reflected objects. 
Most of that noise can be filtered, but silhouettes remain in the map. 
The worst example of such a reflective surface is a~glass wall, which generates a~large amount of noise and also points from both the glass wall and the objects behind. 
The robot can not always detect such a wall and it may try to go through. 

On the other hand, the wall is detected in most cases in a~form of few points which is sufficient for exploration as the points form a~single obstacle avoiding the robot to bump.

The quality and success of the whole exploration highly depends on \textbf{the number of detected visual features}. 
As this number decreases, the SLAM is not able to register the scenes, which leads to misalignment of a~new scan with a map.
Such errors can be sometimes repaired by a~loop closure, but they often remain in the map uncorrected. 


\section*{Acknowledgments}
This work has been supported by the Technology Agency of the Czech Republic under the project no.~TE01020197 \enquote{Centre for Applied Cybernetics} and by the EU Horizon 2020 project~688117 \enquote{SafeLog} .

\bibliographystyle{IEEEtran}
\bibliography{main}

\begin{thebibliography}{1}
\providecommand{\url}[1]{#1}
\csname url@rmstyle\endcsname
\providecommand{\newblock}{\relax}
\providecommand{\bibinfo}[2]{#2}
\providecommand\BIBentrySTDinterwordspacing{\spaceskip=0pt\relax}
\providecommand\BIBentryALTinterwordstretchfactor{4}
\providecommand\BIBentryALTinterwordspacing{\spaceskip=\fontdimen2\font plus
\BIBentryALTinterwordstretchfactor\fontdimen3\font minus
  \fontdimen4\font\relax}
\providecommand\BIBforeignlanguage[2]{{%
\expandafter\ifx\csname l@#1\endcsname\relax
\typeout{** WARNING: IEEEtran.bst: No hyphenation pattern has been}%
\typeout{** loaded for the language `#1'. Using the pattern for}%
\typeout{** the default language instead.}%
\else
\language=\csname l@#1\endcsname
\fi
#2}}

\bibitem{eapd}
\url{http://imr.ciirc.cvut.cz/Research/EAPD}.

\bibitem{icaps}
L.~P. T.~Juchelka, M.~Kulich, ``Multi-robot exploration in the polygonal
  domain,'' in \emph{Int. Conf. on Automated Planning: Workshop on Planning and
  Robotics}, 2013.

\bibitem{rtabmap}
M.~Labbé and F.~Michaud, ``Online global loop closure detection for
  large-scale multi-session graph-based slam,'' in \emph{Intelligent Robots and
  Systems (IROS 2014), 2014 IEEE/RSJ International Conference on}, Sept 2014,
  pp. 2661--2666.

\bibitem{libfreenect}
\url{https://github.com/OpenKinect/libfreenect2}.

\bibitem{bridge}
\url{https://github.com/code-iai/iai\_kinect2/tree/master/kinect2\_bridge}.

\bibitem{rgbd-mapping}
F.~Endres, J.~Hess, J.~Sturm, D.~Cremers, and W.~Burgard, ``3-d mapping with an
  rgb-d camera,'' \emph{IEEE Transactions on Robotics}, vol.~30, no.~1, pp.
  177--187, Feb 2014.

\bibitem{orb-slam}
R.~Mur-Artal, J.~M.~M. Montiel, and J.~D. Tardós, ``{ORB-SLAM}: A versatile
  and accurate monocular {SLAM} system,'' \emph{IEEE Transactions on Robotics},
  vol.~31, no.~5, pp. 1147--1163, Oct 2015.

\end{thebibliography}

\end{document}